\documentclass[letterpaper]{article} 
\usepackage[preprint]{aaai2027}  
\usepackage[hyphens]{url}
\usepackage{graphicx}
\usepackage{natbib}
\usepackage{caption}
\usepackage{latexsym}
\usepackage[utf8]{inputenc}
\usepackage{textcomp}
\usepackage{amsmath}
\usepackage{amssymb}
\usepackage{amsfonts}
\usepackage{bm}
\usepackage{pifont}
\usepackage{algorithm}
\usepackage{algpseudocode}
\usepackage{booktabs}
\usepackage{multirow}
\usepackage{makecell}
\usepackage{tabularx}
\usepackage{arydshln}
\usepackage{array}
\usepackage[most]{tcolorbox}
\usepackage{marvosym}
\definecolor{graybg}{rgb}{0.9,0.9,0.9}
\definecolor{MyDarkBlue}{rgb}{0,0.08,0.45}
\definecolor{MyLightBlue}{RGB}{235,245,255}
\definecolor{ccecolor}{RGB}{0,120,140}
\definecolor{aaecolor}{RGB}{190,90,20}
\definecolor{ttecolor}{RGB}{40,130,60}
\definecolor{onlinecolor}{RGB}{110,50,150}

\newcommand{\gain}[1]{\textcolor[rgb]{0.0, 0.5, 0.0}{$_{\uparrow\textbf{#1}}$}}
\newcommand{\loss}[1]{\textcolor[rgb]{0.8, 0.0, 0.0}{$_{\downarrow\textbf{#1}}$}}
\newcommand{\cmark}{\ding{51}}
\newcommand{\xmark}{\ding{55}}
\newcommand{\method}{SkillCAT}

\title{\method: Contrastive, Assessment-Augmented and Topology-Aware \\ Skill Self-Evolution for LLM Agents}

\author{
    Kunfeng Chen\textsuperscript{\rm 1},
    Qihuang Zhong\textsuperscript{\rm 2},
    Juhua Liu\textsuperscript{\rm 1},
    Bo Du\textsuperscript{\rm 1}
}
\affiliations{
    \textsuperscript{\rm 1}School of Computer Science, Wuhan University, China\\
    \textsuperscript{\rm 2}Nanyang Technological University, Singapore \\
    \{chenkunfeng, liujuhua, dubo\}@whu.edu.cn, qihuang.zhong@ntu.edu.sg 
}

\begin{document}
\maketitle

\begin{abstract}

Skill self-evolution methods for LLM agents aim to turn execution trajectories into reusable skill documents. However, current pipelines typically derive skill patches from a single trajectory per task, merge them indiscriminately, and load the entire skill corpus during inference. These choices lead to unreliable evidence extraction, the accumulation of low-quality or even harmful skill edits, and inefficient use of context due to irrelevant or conflicting skill content. We propose SkillCAT, a framework that decomposes this process into three stages. (1)~Contrastive Causal Extraction (CCE) samples multiple trajectories per task and contrasts same-task success/failure pairs to find the evidence that explains outcome differences. (2)~Assessment-Augmented Evolution (AAE) replays each candidate patch on source-task clones, retains only those that do not damage task outcomes, and then merges the retained patches hierarchically. (3)~Topology-Aware Task Execution (TTE) compiles the evolved skills into routable sub-skill topologies, so that inference loads only task-relevant capability nodes. We evaluate SkillCAT on widely-used agent benchmarks, including SpreadsheetBench, WikiTableQuestions, and DocVQA, and further assess cross-model and out-of-distribution generalization. Across these settings, SkillCAT improves the average score over the initial skill by up to 49.69\%, demonstrating reliable and effective skill evolution.

\end{abstract}

\section{Introduction}
\label{sec:introduction}

\begin{figure}[t]
\centering
\includegraphics[width=\columnwidth]{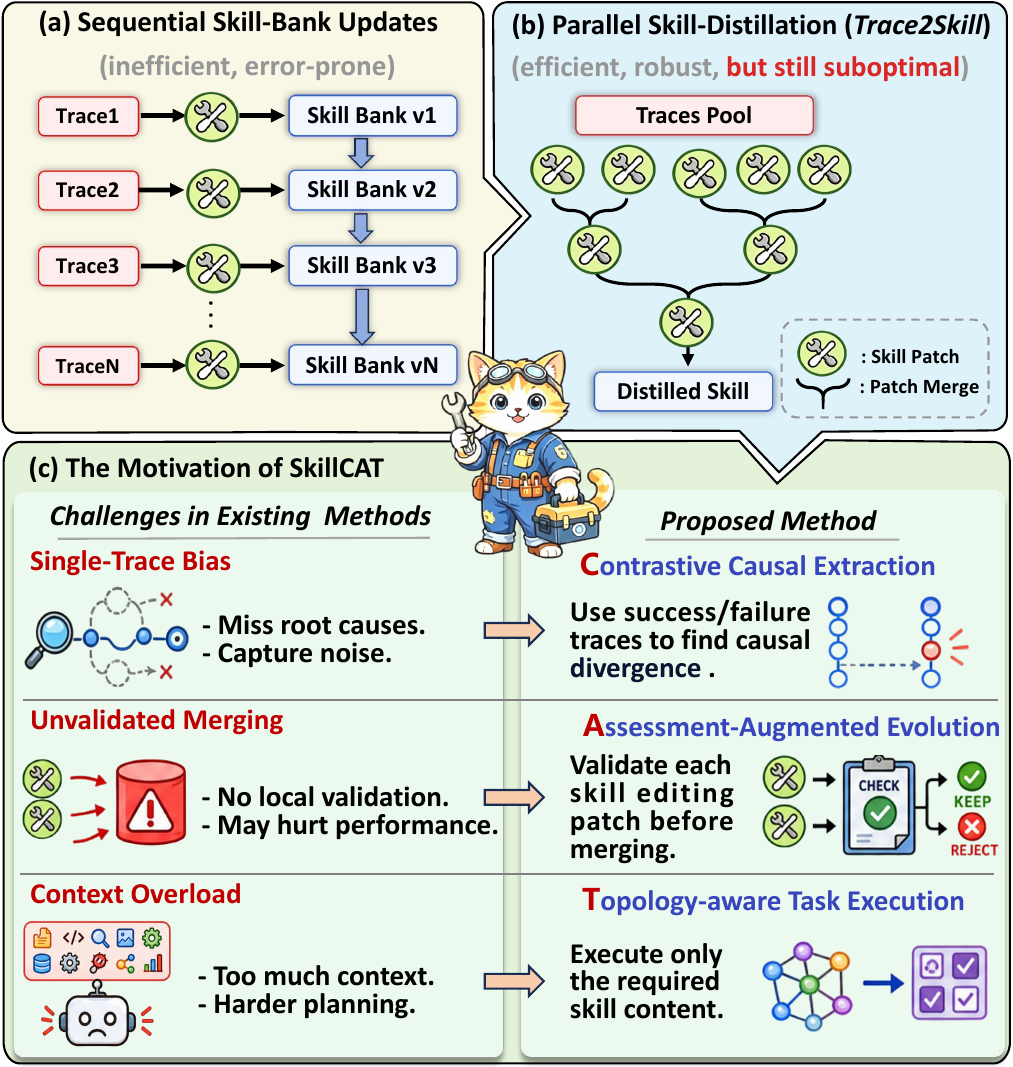}

\caption{\textbf{Three limitations of Trace2Skill-style methods} and the corresponding SkillCAT solutions.}
\label{fig:motivation}
\end{figure}

Large language model (LLM)-based agents have emerged as a general paradigm for solving complex interaction tasks~\cite{wang2024survey,yao2023react,schick2023toolformer,wang2023voyager}. To support this capability,  external skill documents provide an important mechanism for injecting reusable procedures, tool-use knowledge, and task experience at execution time~\cite{xu2026agent,jiang2026sok}. Consequently, with such externalized guiding knowledge, agents can better handle the long-horizon decision processes involved in these tasks without updating model weights. Early skill libraries were mostly hand-written, while recent works enforce the LLM agent to extract and refine skills from its prior trajectories, 
commonly called \emph{skill self-evolution}~\cite{zhou2026comprehensive,yang2026autoskill,zhang2026skillflow}.

Early skill self-evolution methods usually follow a sequential update loop: each new trajectory edits the current skill once, and the pipeline then moves on to the next trajectory~\cite{shinn2023reflexion,madaan2023self,zhang2026memskill,jiang2026xskill,chen2026skillcraft}. This procedure is simple, but later edits tend to yield diminishing returns, accumulating noise or overwriting useful behavior. Trace2Skill~\cite{ni2026trace2skill} recasts this process as an offline batch pipeline: it extracts one skill patch from each trajectory and then consolidates these patches into a more general skill edit. Building on this paradigm, SkillOpt~\cite{yang2026skillopt} models skill self-evolution as gradient descent and applies iterative optimization to further refine it. Overall, these methods establish a paradigm in which agents' execution histories are distilled into reusable skill updates.

Despite these advances, Trace2Skill-style pipelines still face three challenges as task complexity increases and the skill context grows. \textbf{(1) Single-Trace Bias}: a single trajectory often provides weak evidence for a task, \textit{i.e.}, a successful trace may reflect an accidental strategy, while a failed trace rarely reveals the root cause~\cite{ni2026trace2skill,li2026skillsbench}. \textbf{(2) Unvalidated Merging}: patches are merged without independently verifying whether they benefit the source task, so low-quality or even harmful patches may enter the skill corpus~\cite{zhang2026evoskills,tian2026skills,gou2024critic}. \textbf{(3) Context Overload}: irrelevant or conflicting rules are fed into the agent during inference, increasing inference overhead and degrading performance~\cite{li2026graph,meng2026skillrae,chen2026try}.

In response to these problems, we propose a \textbf{C}ontrastive, \textbf{A}ssessment-Augmented, and \textbf{T}opology-Aware skill self-evolution (namely \textbf{SkillCAT}), which decomposes the skill lifecycle into three observable stages and introduces targeted optimizations for each. \textbf{Contrastive Causal Extraction (CCE)} generates multiple trajectories per task via multi-seed sampling and builds same-task success/failure contrastive pairs, extracting candidate experience around the causal watershed between outcomes rather than summarizing the full trace. \textbf{Assessment-Augmented Evolution (AAE)} replays each candidate patch on source-task clones, assigns calibrated scores based on outcome transitions, and retains only patches that improve or preserve source-task behavior for hierarchical merging. \textbf{Topology-Aware Task Execution (TTE)} compiles the evolved skill into a routable topology of capability nodes and loads only task-relevant content during inference. In this way, SkillCAT addresses the three challenges by grounding skill edits in stronger contrastive evidence, filtering them through task-level validation, and exposing only relevant skill content during execution.

Empirically, we evaluate SkillCAT under both same-model and cross-model skill-use settings. In the same-model setting, Qwen3.5-35B-A3B and Qwen3.5-122B-A10B each act as both skill author and skill user on agent-assisted office tasks (\textit{e.g.}, spreadsheet manipulation), and the evolved skills are further tested on out-of-distribution (OOD) table question answering. We then evaluate transfer to unseen skill users, including Gemma-4-31B-it and GPT-5.4-mini, and assess multimodal document question answering with Qwen-authored skills reused by two Qwen users. Across these settings, SkillCAT improves the average score over the initial skill by up to 49.69\%. These results indicate that SkillCAT remains robust across weaker skill initialization, unseen task domains, and different user models, and our ablation studies further confirm that every component (CCE, AAE, and TTE) contributes to the final improvement.

We summarize our contributions as follows:
\begin{itemize}

\item We identify three practical problems that limit current skill self-evolution: single-trace bias, unvalidated merging, and inference-time context overload, which together lead to suboptimal performance and inefficient inference.
\item We propose SkillCAT, which addresses these problems with three simple-yet-effective components: same-task success/failure contrastive extraction, source-task replay before merging, and topology-aware routing.

\item Extensive experiments on four LLM agents and three agentic benchmarks show that SkillCAT improves the average score over matched initial-skill baselines by up to 49.69\%, demonstrating its effectiveness and generality.
\end{itemize}

\begin{figure*}[!t]
\centering
\includegraphics[width=\textwidth]{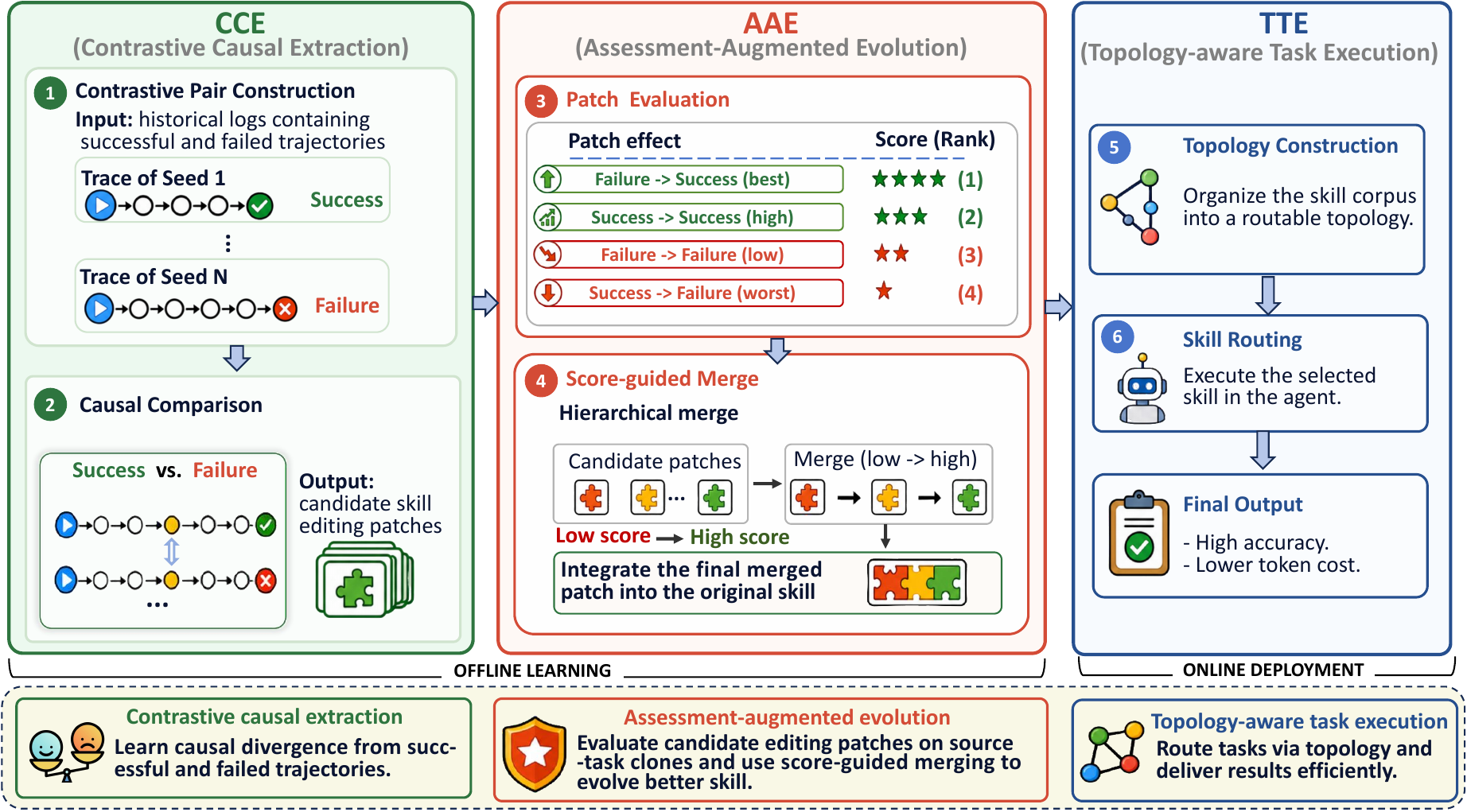}

\caption{\textbf{Overview of SkillCAT pipeline}, comprising three key components: (1) CCE extracts same-task contrastive evidence by comparing successful and failed trajectories, (2) AAE validates candidate patches and merges only the helpful ones into the skill, and (3) TTE organizes the skill into a routable topology and retrieves task-relevant nodes for agent execution.}
\label{fig:framework}
\end{figure*}

\section{Related Work}
\label{sec:related_work}
\paragraph{Skill for LLM Agents.}
Agent skills encode reusable task workflows, tool-use strategies, and execution constraints as external artifacts, enabling task-specific guidance~\cite{wang2023voyager,zhou2026comprehensive}. Such skills may be human-written or model-generated, yet their presence alone does not ensure better task performance. SkillsBench~\cite{li2026skillsbench} finds that curated skills are generally helpful but automatically generated skills yield inconsistent gains. SkillLearnBench~\cite{zhong2026skilllearnbench} and SkillFlow~\cite{zhang2026skillflow} further show that continual updates can drift or regress. A separate line of work studies how agents select and compose skills from large collections, such as Graph of Skills~\cite{li2026graph}, SkillRAE~\cite{meng2026skillrae}, and GraSP~\cite{xia2026grasp}.
These methods primarily address inter-skill retrieval and composition, typically assuming that useful skill artifacts already exist. Our setting is complementary: rather than expanding a skill library, we refine a single base skill into a more reliable artifact and expose only task-relevant parts of that same skill at test time.

\paragraph{Skill Self-Evolution.}
Skill self-evolution turns execution experience into persistent updates to external skills without modifying the backbone model~\cite{zhou2026comprehensive}. For instance, AutoSkill~\cite{yang2026autoskill}, XSkill~\cite{jiang2026xskill}, and SkillClaw~\cite{ma2026skillclaw} extend 
this process to personalized, multimodal, and cross-user experience, respectively. Within execution-driven skill evolution, representative methods adopt different update strategies, including consolidating trajectory patches, optimizing a single skill document, and iteratively expanding and selecting a skill set. 
Specifically, Trace2Skill~\cite{ni2026trace2skill} starts from an initial skill, extracts trajectory-local patches in parallel from a fixed trace pool, and hierarchically consolidates them into a single skill directory. SkillOpt~\cite{yang2026skillopt} treats a single skill document as trainable external state, converts scored rollouts into bounded document edits, and accepts only updates that improve held-out validation performance. EvoSkill~\cite{alzubi2026evoskill} follows failure-driven iterative evolution, creating or revising skills from execution failures and using held-out validation to select promising candidate agent configurations. 
Despite their effectiveness, these methods still neglect three practical problems:
evidence drawn from isolated executions, patches integrated without checking their task-level effect, and indiscriminate full-skill loading. 
In response to these problems, SkillCAT systematically integrates same-task contrastive extraction, source-task replay validation, and topology-aware routing for a single evolving skill, enabling reliable refinement and task-selective deployment.

\section{Method}
\label{sec:method}

In this section, we introduce SkillCAT, which decomposes skill self-evolution into three stages: Contrastive Causal Extraction (CCE), Assessment-Augmented Evolution (AAE), and Topology-Aware Task Execution (TTE). Notably, CCE and AAE operate offline during skill learning, while TTE operates online during task deployment.

\subsection{Problem Formulation}
\label{sec:problem_formulation}

We consider a skill self-evolution setting in which an agent is given a base skill $S_0$, a set of evolution tasks $\mathcal{X} = \{x_1, \ldots, x_N\}$ and a set of test tasks $\mathcal{X^*} = \{x^*_1, \ldots, x^*_M\}$. By executing each task multiple times, we collect diverse execution traces, each labeled as success or failure by the official evaluator. In SkillCAT, the offline stage outputs an evolved skill $S^*$, and the online execution stage assembles a routed skill $S_j$ from $S^*$ for each test task $x^*_j \in \mathcal{X^*}$. The objective of SkillCAT is to improve execution performance on unseen tasks while keeping the skill content injected into the agent context compact and task-relevant.

Therefore, our goal is not merely to summarize all past execution traces into experience patches and integrate them into $S_0$ to produce a comprehensive skill document, but to make skill evolution \emph{selective}. To this end, SkillCAT involves (1) identifying reliable execution evidence, (2) validating which candidate patches should be incorporated into the evolved skill, and (3) exposing only task-relevant skill content during execution. Figure~\ref{fig:framework} and Algorithm~\ref{alg:skillcat} summarize the complete pipeline. CCE extracts reliable candidate experience from multi-seed trajectories, AAE filters and merges candidate patches via source-task replay, and TTE assembles a task-relevant skill with a routable topology. These three modules correspond to evidence acquisition, patch validation and integration, and test-time skill deployment, respectively.

\subsection{Contrastive Causal Extraction (CCE)}
\label{sec:cce}

CCE forms same-task success/failure pairs from multi-seed runs and extracts skill evidence at the meaningful divergence between them. Since each pair of samples shares the same input, tools, and evaluator, the extracted evidence can be used to explain why two different execution choices lead to different execution results. 

\paragraph{Contrastive pair construction.}

For task $x_i$, we conduct multiple runs with different random seeds. Let $\mathcal{T}_i^+$ and $\mathcal{T}_i^-$ denote the sets of successful and failed trajectories, respectively. When both sets are non-empty, CCE randomly samples one trajectory from each to form a contrastive pair $(\tau_i^+, \tau_i^-)$. This same-task pairing  isolates the critical behaviors that distinguish success from failure.

\paragraph{Causal comparison.}

Given a contrastive pair $(\tau_i^+, \tau_i^-)$, CCE guides an LLM-based contrastive extractor $E$ to identify the point where the action sequences in the successful and failed traces diverge. Then $E$ uses this divergence as an internal reasoning clue to write a candidate experience record $r_i$, including local evidence, the cause of failure and a skill-editable lesson. This process can be formulated as:

\begin{equation}
\label{eq:record_extraction}
r_i = E(\tau_i^+, \tau_i^-).
\end{equation}

Unlike existing methods that summarize the full trace~\cite{ni2026trace2skill,alzubi2026evoskill}, our candidate experience record focuses on the behavioral difference where outcomes diverge. For tasks lacking contrastive outcomes (\textit{i.e.}, pure success or failure), CCE retains a single-trajectory pipeline to extract non-contrastive records. Finally, a skill editor combines the extracted record $r_i$ with the base skill $S_0$ to generate candidate skill editing patch $p_i$, which is then passed to AAE for validation and merging.

\subsection{Assessment-Augmented Evolution (AAE)}
\label{sec:aae}

AAE treats each candidate patch as a hypothesis to be verified rather than a rule to be merged directly. It isolates the patch in a temporary skill, replays it on source-task, and scores the resulting outcome transition before skill merging.

\paragraph{Patch evaluation.}

Given a candidate patch $p_i$, AAE replays the temporary skill induced by $p_i$ on the associated source-task clones and records both the original and replay outcomes. For simplicity, let $y_i, \hat{y}_i \in \{0,1\}$ denote the original and replay outcomes, respectively, where $1$ denotes success and $0$ denotes failure. AAE ranks the four outcome transitions directly:

\begin{equation}
\label{eq:aae_scoring}
a_i = \begin{cases}
3.0 & (y_i,\hat{y}_i)=(0,1), \\
2.0 & (y_i,\hat{y}_i)=(1,1), \\
1.0 & (y_i,\hat{y}_i)=(0,0), \\
0.0 & (y_i,\hat{y}_i)=(1,0).
\end{cases}
\end{equation}

This scoring rule ranks candidate patches by how they change the source-task outcome. The transition of (\texttt{failure}$\to$\texttt{success}) receives the highest score, as the patch repairs the source task, whereas (\texttt{success}$\to$\texttt{success}) is accepted as behavior preservation. The transition of (\texttt{failure}$\to$\texttt{failure}) is downweighted, and (\texttt{success}$\to$\texttt{failure}) is rejected.

\paragraph{Score-guided merge.}
We only merge patches whose scores are greater than the threshold $\theta$:
\begin{equation}
\label{eq:patch_selection}
\mathcal{P}_{\theta} = \{ p_i : a_i \geq \theta \},
\end{equation}
where $\theta = 2.0$ in our work. Therefore, the retained patches either repair a source-task failure or preserve a known success, whereas patches that leave failures unresolved or turn successes into failures are excluded from the global skill.

After thresholding, AAE groups the retained patches into score tiers and merges them from the lowest tier to the highest. Let $\ell$ index the merge tier, $L$ denote the number of tiers, and $\mathcal{P}_{\theta}^{(\ell)}$ denote the selected patches at tier $\ell$. Let $\mu$ denote the skill-merge operator:
\begin{equation}
\label{eq:tiered_merge}
S^{(\ell)} = \mu\!\left(S^{(\ell-1)},\; \mathcal{P}_{\theta}^{(\ell)}\right),
\end{equation}

where $S^{(0)} = S_0$ and the final evolved skill is $S^* = S^{(L)}$. This ordering gives higher-scoring patches priority in later stages, while still letting lower-scoring, validated ones contribute their non-conflicting rules. Besides patch selection and merge order, AAE directly reuses Trace2Skill’s edit procedure, extracting general principles from recurrent edits rather than memorizing instance-specific fixes.

\begin{algorithm}[t]
\caption{SkillCAT Pipeline}
\label{alg:skillcat}
\small
\begin{algorithmic}[1]
\State \textbf{Input:} Tasks for evolution $\mathcal{X}$, tasks for test $\mathcal{X^*}$, base skill $S_0$, score threshold $\theta$, node budget $k$.
\State \textbf{Output:} Evolved skill $S^*$, skill topology $(S_c, \mathcal{V}, \mathcal{G})$, and routed skill $S_j$ for each test task.

\Statex \colorbox{gray!15}{\parbox{\dimexpr\linewidth-2\fboxsep}{\textbf{Stage 1: Contrastive Causal Extraction}}}

\For{each task $x_i \in \mathcal{X}$ ($i = 1, \dots, N$)}
    \State Run the agent with multiple random seeds to get traces $\mathcal{T}_i$.
    \State Split $\mathcal{T}_i$ into successful traces $\mathcal{T}_i^+$ and failed traces $\mathcal{T}_i^-$.
    \If{$\mathcal{T}_i^+ \neq \emptyset$ \textbf{and} $\mathcal{T}_i^- \neq \emptyset$}
        \State Sample a contrastive pair $(\tau_i^+, \tau_i^-)$ from $\mathcal{T}_i^+$ and $\mathcal{T}_i^-$.
        \State Identify divergence and extract record 
        \State $r_i \gets E(\tau_i^+, \tau_i^-)$.
    \Else
        \State Extract a non-contrastive record $r_i$ from $\mathcal{T}_i$.
    \EndIf
    \State Generate candidate skill editing patch $p_i$ from $r_i$ and $S_0$.
\EndFor

\Statex \colorbox{gray!15}{\parbox{\dimexpr\linewidth-2\fboxsep}{\textbf{Stage 2: Assessment-Augmented Evolution }}}
\State $\mathcal{P}_\theta \gets \emptyset$
\For{each patch $p_i$ ($i = 1, \dots, N$)}
    \State Load the skill built from $p_i$ and rerun the  source-task.
    \State Compute assessment score $a_i$ via Eq.~\ref{eq:aae_scoring}.
    \State $\mathcal{P}_\theta \gets \{p_i : a_i \geq \theta\}$ \Comment{Score-thresholded selection}.
\EndFor
\State Merge $\mathcal{P}_\theta$ sequentially using $\mu$ to obtain $S^*$ via Eq.~\ref{eq:tiered_merge}.

\Statex \colorbox{gray!15}{\parbox{\dimexpr\linewidth-2\fboxsep}{\textbf{Stage 3: Topology-Aware Task Execution }}}
\State Compile $S^*$ into $(S_c, \mathcal{V}, \mathcal{G})$ and node bodies $\{B_v\}_{v \in \mathcal{V}}$.
\For{each task $x^*_j \in \mathcal{X^*}$ ($j = 1, \dots, M$)} \Comment{Online}
    \State $\mathcal{V}_j \gets R(x^*_j, \mathcal{G}, k)$ \Comment{LLM routing, Eq.~\ref{eq:tte_routing}}.
    \State $S_j \gets A\!\left(S_c, \{B_v\}_{v \in \mathcal{V}_j} \right)$ \Comment{LLM assembly, Eq.~\ref{eq:routed_skill_assembly}}.
\EndFor
\end{algorithmic}
\end{algorithm}

\begin{table*}[!t]
\centering
\fontsize{9pt}{11pt}\selectfont
\setlength{\tabcolsep}{3pt}

\newlength{\numwidth}
\settowidth{\numwidth}{\textbf{82.85\gain{59.12}}}
\newcommand{\lc}[1]{\makebox[\numwidth][l]{#1}}

\begin{tabular}{@{} >{\raggedright\arraybackslash}p{0.09\textwidth} l *{7}{c} @{}}
\toprule
\multirow{2}{*}{\textbf{Skill User}} & \multirow{2}{*}{\textbf{Method}}
  & \multicolumn{3}{c}{\textbf{Skill Author: Qwen3.5-35B-A3B}}
  & \multicolumn{3}{c}{\textbf{Skill Author: Qwen3.5-122B-A10B}}
  & \multirow{2}{*}{\textbf{Overall}} \\
\cmidrule(lr){3-5} \cmidrule(lr){6-8}
  & 
  & \textbf{SpreadsheetBench} & \textbf{WikiTQ } & \textbf{Avg.}
  & \textbf{SpreadsheetBench} & \textbf{WikiTQ } & \textbf{Avg.} & \\
\midrule
\multirow{11}{*}{\shortstack[l]{\textbf{Qwen3.5-}\\\textbf{35B-A3B}}}
  & No-Skill & \lc{19.00} & \lc{13.33} & \lc{16.17} & \lc{19.00} & \lc{13.33} & \lc{16.17} & \lc{16.17} \\
  \cdashline{2-9}
  & Human-Written & \lc{9.67} & \lc{9.02} & \lc{9.35} & \lc{9.67} & \lc{9.02} & \lc{9.35} & \lc{9.35} \\
  & \hspace{1em}- Trace2Skill & \lc{29.67\gain{20.00}} & \lc{51.22\gain{42.20}} & \lc{40.45\gain{31.10}} & \lc{30.83\gain{21.16}} & \lc{15.66\gain{6.64}} & \lc{23.25\gain{13.90}} & \lc{31.85\gain{22.50}} \\
  & \hspace{1em}- EvoSkill & \lc{37.83\gain{28.16}} & \lc{48.87\gain{39.85}} & \lc{43.35\gain{34.00}} & \lc{25.83\gain{16.16}} & \lc{56.31\gain{47.29}} & \lc{41.07\gain{31.72}} & \lc{42.21\gain{32.86}} \\
  & \hspace{1em}- SkillOpt & \lc{33.33\gain{23.66}} & \lc{48.54\gain{39.52}} & \lc{40.94\gain{31.59}} & \lc{34.50\gain{24.83}} & \lc{40.78\gain{31.76}} & \lc{37.64\gain{28.29}} & \lc{39.29\gain{29.94}} \\
  & \hspace{1em}- \textbf{SkillCAT (Ours)} & \lc{\textbf{55.00}\gain{45.33}} & \lc{\textbf{78.64}\gain{69.62}} & \lc{\textbf{66.82}\gain{57.47}} & \lc{\textbf{37.50}\gain{27.83}} & \lc{\textbf{65.00}\gain{55.98}} & \lc{\textbf{51.25}\gain{41.90}} & \lc{\textbf{59.04}\gain{49.69}} \\
\cmidrule{2-9}
  & LLM-Gen & \lc{20.17} & \lc{20.14} & \lc{20.16} & \lc{20.17} & \lc{20.14} & \lc{20.16} & \lc{20.16} \\
  & \hspace{1em}- Trace2Skill & \lc{20.00\loss{0.17}} & \lc{38.14\gain{18.00}} & \lc{29.07\gain{8.91}} & \lc{19.00\loss{1.17}} & \lc{49.84\gain{29.70}} & \lc{34.42\gain{14.26}} & \lc{31.75\gain{11.59}} \\
  & \hspace{1em}- EvoSkill & \lc{51.67\gain{31.50}} & \lc{64.56\gain{44.42}} & \lc{58.11\gain{37.95}} & \lc{29.17\gain{9.00}} & \lc{\textbf{59.87}\gain{39.73}} & \lc{44.52\gain{24.36}} & \lc{51.32\gain{31.16}} \\
  & \hspace{1em}- SkillOpt & \lc{32.00\gain{11.83}} & \lc{51.46\gain{31.32}} & \lc{41.73\gain{21.57}} & \lc{33.17\gain{13.00}} & \lc{54.69\gain{34.55}} & \lc{43.93\gain{23.77}} & \lc{42.83\gain{22.67}} \\
  & \hspace{1em}- \textbf{SkillCAT (Ours)} & \lc{\textbf{54.00}\gain{33.83}} & \lc{\textbf{71.36}\gain{51.22}} & \lc{\textbf{62.68}\gain{42.52}} & \lc{\textbf{37.17}\gain{17.00}} & \lc{54.37\gain{34.23}} & \lc{\textbf{45.77}\gain{25.61}} & \lc{\textbf{54.23}\gain{34.07}} \\
\midrule
\multirow{11}{*}{\shortstack[l]{\textbf{Qwen3.5-}\\\textbf{122B-A10B}}}
  & No-Skill & \lc{27.67} & \lc{21.50} & \lc{24.59} & \lc{27.67} & \lc{21.50} & \lc{24.59} & \lc{24.59} \\
  \cdashline{2-9}
  & Human-Written & \lc{48.33} & \lc{74.68} & \lc{61.51} & \lc{48.33} & \lc{74.68} & \lc{61.51} & \lc{61.51} \\
  & \hspace{1em}- Trace2Skill & \lc{55.00\gain{6.67}} & \lc{77.33\gain{2.65}} & \lc{66.17\gain{4.66}} & \lc{\textbf{69.83}\gain{21.50}} & \lc{79.24\gain{4.56}} & \lc{74.54\gain{13.03}} & \lc{70.36\gain{8.85}} \\
  & \hspace{1em}- EvoSkill & \lc{55.83\gain{7.50}} & \lc{73.14\loss{1.54}} & \lc{64.49\gain{2.98}} & \lc{53.17\gain{4.84}} & \lc{81.23\gain{6.55}} & \lc{67.20\gain{5.69}} & \lc{65.85\gain{4.34}} \\
  & \hspace{1em}- SkillOpt & \lc{\textbf{62.17}\gain{13.84}} & \lc{80.26\gain{5.58}} & \lc{\textbf{71.21}\gain{9.70}} & \lc{54.00\gain{5.67}} & \lc{79.94\gain{5.26}} & \lc{66.97\gain{5.46}} & \lc{69.09\gain{7.58}} \\
  & \hspace{1em}- \textbf{SkillCAT (Ours)} & \lc{59.50\gain{11.17}} & \lc{\textbf{81.55}\gain{6.87}} & \lc{70.53\gain{9.02}} & \lc{69.17\gain{20.84}} & \lc{\textbf{81.55}\gain{6.87}} & \lc{\textbf{75.36}\gain{13.85}} & \lc{\textbf{72.95}\gain{11.44}} \\
\cmidrule{2-9}
  & LLM-Gen & \lc{26.17} & \lc{23.73} & \lc{24.95} & \lc{26.17} & \lc{23.73} & \lc{24.95} & \lc{24.95} \\
  & \hspace{1em}- Trace2Skill & \lc{25.33\loss{0.84}} & \lc{54.55\gain{30.82}} & \lc{39.94\gain{14.99}} & \lc{26.33\gain{0.16}} & \lc{56.05\gain{32.32}} & \lc{41.19\gain{16.24}} & \lc{40.57\gain{15.62}} \\
  & \hspace{1em}- EvoSkill & \lc{63.33\gain{37.16}} & \lc{78.32\gain{54.59}} & \lc{70.83\gain{45.88}} & \lc{62.50\gain{36.33}} & \lc{81.88\gain{58.15}} & \lc{72.19\gain{47.24}} & \lc{71.51\gain{46.56}} \\
  & \hspace{1em}- SkillOpt & \lc{68.00\gain{41.83}} & \lc{81.55\gain{57.82}} & \lc{74.78\gain{49.83}} & \lc{61.50\gain{35.33}} & \lc{78.64\gain{54.91}} & \lc{70.07\gain{45.12}} & \lc{72.43\gain{47.48}} \\
  & \hspace{1em}- \textbf{SkillCAT (Ours)} & \lc{\textbf{69.67}\gain{43.50}} & \lc{\textbf{82.85}\gain{59.12}} & \lc{\textbf{76.26}\gain{51.31}} & \lc{\textbf{63.83}\gain{37.66}} & \lc{\textbf{82.20}\gain{58.47}} & \lc{\textbf{73.02}\gain{48.07}} & \lc{\textbf{74.64}\gain{49.69}} \\
\midrule
\multicolumn{9}{c}{\textit{Cross-Family Model Generalization}} \\
\midrule
\multirow{3}{*}{\shortstack[l]{\textbf{Gemma-4-}\\\textbf{31B-it}}}
  & No-Skill & \lc{46.17} & \lc{\textbf{77.99}} & \lc{62.08} & \lc{46.17} & \lc{\textbf{77.99}} & \lc{62.08} & \lc{62.08} \\
  & Human-Written & \lc{39.83} & \lc{76.70} & \lc{58.27} & \lc{39.83} & \lc{76.70} & \lc{58.27} & \lc{58.27} \\
  & \hspace{1em}- \textbf{SkillCAT (Ours)} & \lc{\textbf{61.33}\gain{21.50}} & \lc{77.67\gain{0.97}} & \lc{\textbf{69.50}\gain{11.23}} & \lc{\textbf{70.00}\gain{30.17}} & \lc{75.08\loss{1.62}} & \lc{\textbf{72.54}\gain{14.27}} & \lc{\textbf{71.02}\gain{12.75}} \\
\cmidrule{2-9}
\multirow{3}{*}{\shortstack[l]{\textbf{GPT-5.4-}\\\textbf{mini}}}
  & No-Skill & \lc{10.50} & \lc{44.34} & \lc{27.42} & \lc{10.50} & \lc{44.34} & \lc{27.42} & \lc{27.42} \\
  & Human-Written & \lc{31.00} & \lc{51.78} & \lc{41.39} & \lc{31.00} & \lc{51.78} & \lc{41.39} & \lc{41.39} \\
  & \hspace{1em}- \textbf{SkillCAT (Ours)} & \lc{\textbf{37.50}\gain{6.50}} & \lc{\textbf{59.55}\gain{7.77}} & \lc{\textbf{48.53}\gain{7.14}} & \lc{\textbf{32.50}\gain{1.50}} & \lc{\textbf{64.72}\gain{12.94}} & \lc{\textbf{48.61}\gain{7.22}} & \lc{\textbf{48.57}\gain{7.18}} \\
\bottomrule
\end{tabular}
\caption{
    \textbf{Main results on SpreadsheetBench and WikiTableQuestions (WikiTQ).}
    SpreadsheetBench evaluates held-out accuracy, while WikiTQ assesses OOD generalization. Our results are averaged over three random seeds to minimize uncertainty. Subscript arrows indicate performance changes relative to the corresponding initial-skill baselines.  Best results are in \textbf{bold}. 
}
\label{tab:main_results}
\end{table*}

\subsection{Topology-Aware Task Execution (TTE)}
\label{sec:tte}

TTE casts test-time skill use as a context-selection problem. It compiles the evolved skill $S^*$ into a routable topology and assembles a compact routed skill $S_j$ for each task, avoiding the cost and distraction of loading the full skill.

\paragraph{Topology construction.}

TTE compiles the AAE-generated skill $S^*$ into a core skill $S_c$ and a set of capability nodes $\mathcal{V}$. For each node, TTE keeps the original body $B_v$ for later assembly and extracts its title, keywords, summary, and dependencies as routing metadata. The node metadata and dependency edges form a compact topology summary $\mathcal{G}$, in which edges denote procedural or tool-use dependencies between nodes. In this way, routing reads only $\mathcal{G}$, whereas assembly accesses the original bodies of the selected nodes.

\paragraph{Skill routing.}
Let $R$ denote the LLM router and $A$ denote the LLM skill assembler. Given a test task $x^*_j$ and a node budget $k$, $R$ selects relevant capability nodes from the topology summary $\mathcal{G}$.
\begin{equation}
\label{eq:tte_routing}
\mathcal{V}_j = R(x^*_j, \mathcal{G}, k), \qquad |\mathcal{V}_j| \le k .
\end{equation}
The assembler $A$ then constructs the runtime skill $S_j$ from the core skill and the selected node bodies.
\begin{equation}
\label{eq:routed_skill_assembly}
S_j = A\!\left(S_c, \{B_v\}_{v \in \mathcal{V}_j}\right).
\end{equation}


\section{Experiments}
\label{sec:experiments}

\subsection{Experimental Setup}
\label{sec:experimental_setup}

\paragraph{Datasets and Evaluation.}

Our primary benchmark for evaluating domain-specific agent skills is SpreadsheetBench-Verified, a human-validated subset of SpreadsheetBench~\cite{spreadsheetbench2024}. It contains real-world Excel forum questions covering cell- and sheet-level spreadsheet manipulation. Following Trace2Skill~\cite{ni2026trace2skill} and SkillOpt~\cite{yang2026skillopt}, we split its 400 samples into 200 for evolution and the other 200 for held-out testing.
The protocol executes the solution on the input workbook and compares output cells against the gold workbook, scoring a task as correct only when all answers match. To test out-of-distribution (OOD) generalization, we also evaluate on WikiTableQuestions~\cite{pasupat2015wikitq} (WikiTQ), a semi-structured table QA benchmark over Wikipedia tables, comparing predicted and gold answer denotations under the official protocol and reporting accuracy. 
To test multimodal generalization, we evaluate on DocVQA~\cite{mathew2021docvqa} using its official validation split, which contains 5,349 question–image pairs. The first 2,700 question-image pairs are used for evolution and the remaining 2,649 are for evaluation, and we report the ANLS~\cite{mathew2021docvqa} and Acc (ANLS $\geq$ 0.5) results. To minimize uncertainty, our results are averaged over 3 random seeds.

\paragraph{Baseline and Models.}

Following~\citet{ni2026trace2skill}, in addition to the \textit{No-Skill} baseline, we evaluate our SkillCAT under two skill-based settings: 1) \textit{Human-Written}, which starts from Anthropic's official \texttt{xlsx} skill; and 2) \textit{LLM-Gen}, which starts from a skill generated by the corresponding LLM-based skill authors. In both settings, we compare SkillCAT against three cutting-edge counterparts: \textit{Trace2Skill}~\cite{ni2026trace2skill}, \textit{EvoSkill}~\cite{alzubi2026evoskill}, and \textit{SkillOpt}~\cite{yang2026skillopt}. The full details of all baselines are provided in the supplementary material. Similar to Trace2Skill, we employ Qwen3.5-35B-A3B and Qwen3.5-122B-A10B~\cite{qwen3.5} as both skill authors and skill users. Moreover, for cross-model evaluation, we reuse the skills evolved by the Qwen models under the \textit{Human-Written} setting and evaluate them on two additional models, \textit{i.e.}, Gemma-4-31B-it~\cite{gemmateam2026gemma4} and GPT-5.4-mini~\cite{singh2025openai}.

\paragraph{Implementation Details.}

All agents run in a ReAct-style~\cite{yao2023react} harness with filesystem and spreadsheet tools. Specifically, in SkillCAT, CCE samples 5 trajectory seeds per evolution task to collect success/failure evidence. AAE evaluates candidate patches on source-task clones and keeps only patches with $a_i \geq 2.0$ before hierarchical merging, and TTE uses the graph-based router with \texttt{Top-k}=7, followed by task-specific skill injection. 

\begin{figure}[t]
\centering
\includegraphics[width=\columnwidth]{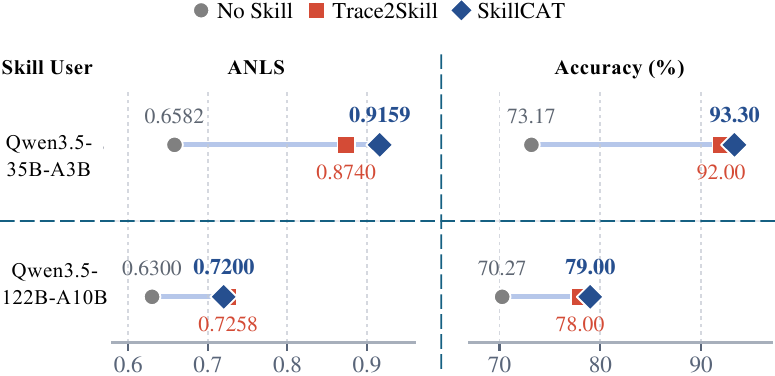}
\caption{\textbf{Multimodal evaluation on DocVQA}. Skills authored by Qwen3.5-35B-A3B are tested with matched and Qwen3.5-122B-A10B users.}
\label{fig:docvqa}
\end{figure}

\subsection{Main Results}
\label{sec:main_results}

\paragraph{SkillCAT improves average performance across in-domain and OOD settings.}

The main results in Table~\ref{tab:main_results} show that SkillCAT achieves the highest overall average for all four combinations of Qwen skill user and skill initialization, outperforming all initialization and skill-evolution baselines.
Its largest margin over the strongest skill-evolution baseline occurs for the Qwen3.5-35B-A3B user with Human-Written initialization: when averaged across the two skill authors and both tasks, SkillCAT reaches 59.04\%, compared with 42.21\% for the second-best (EvoSkill), a margin of 16.83\%. In the stronger Human-Written setting with Qwen3.5-122B-A10B as both author and user, the strongest competing method differs by task: Trace2Skill leads SkillCAT by 0.66\% on SpreadsheetBench, whereas SkillCAT leads the strongest WikiTQ competitor, EvoSkill, by 0.32\%. Despite these task-level differences, SkillCAT retains the highest aggregate score at 72.95\%. These small task-level differences may occur because the strong initial skill already performs well, leaving relatively little room for improvement.

\paragraph{SkillCAT skills show average gains on unseen user models without re-evolution.}

The cross-model evaluation further tests whether the improvements stem from reusable skill content rather than author-specific behavior. 
As shown in Table~\ref{tab:main_results}, skills produced by either Qwen author improve the average score of both unseen users without re-evolution: by 11.23\% and 14.27\% for Gemma-4-31B-it and by 7.14 and 7.22 points for GPT-5.4-mini. However, task-level gains appear to depend on the recipient's baseline performance. On SpreadsheetBench, where Gemma scores 46.17\% without skills, SkillCAT gains 21.50 and 30.17 points over Human-Written. On WikiTQ, Gemma's \textit{No-Skill} score is already 77.99\%, while the two variants reach 77.67\% and 75.08\%, or $+0.97$ and $-1.62$ points relative to Human-Written. This suggests that the benefits of cross-model skills maybe relatively small when the recipient has already performed strongly in the task.

\paragraph{SkillCAT achieves consistent gains in multimodal settings.}
The multimodal evaluation provides further evidence for the effectiveness of SkillCAT and underscores the benefits of selectively merging patches. As shown in Figure~\ref{fig:docvqa}, SkillCAT consistently achieves substantial gains over \textit{No-Skill} on both user model scales.
Compared to Trace2Skill, SkillCAT also demonstrates a clear advantage. 
On the matched Qwen3.5-35B-A3B user, SkillCAT outperforms Trace2Skill in both metrics. When applied to the larger Qwen3.5-122B-A10B user, although both methods yield comparable ANLS, SkillCAT still ensures higher Accuracy. This consistency suggests that validating candidate edits helps prevent noisy, approximate matches and instead yields higher-quality, more precise generation behavior.

\begin{table}[t]

\centering
\fontsize{9pt}{10pt}\selectfont
\setlength{\tabcolsep}{9pt}

\begin{tabular}{l ccc c}
\toprule
\multirow{2}{*}{\textbf{Condition}} & \multicolumn{3}{c}{\textbf{Modules}} & \textbf{Acc.} \\
\cmidrule(lr){2-4} \cmidrule(lr){5-5}
 & \small CCE & \small AAE & \small TTE & (\%) $\uparrow$ \\
\midrule
Trace2Skill & \xmark & \xmark & \xmark & 29.67 \\
\textbf{SkillCAT Full} & \cmark & \cmark & \cmark & \textbf{55.00}\gain{25.33} \\
\midrule
\textit{-w/o} CCE & \xmark & \cmark & \cmark & 32.50\gain{2.83} \\
\textit{-w/o} AAE & \cmark & \xmark & \cmark & 26.00\loss{3.67} \\
\textit{-w/o} TTE & \cmark & \cmark & \xmark & 46.50\gain{16.83} \\
\hdashline
Only CCE & \cmark & \xmark & \xmark & 39.00\gain{9.33} \\
Only AAE & \xmark & \cmark & \xmark & 34.00\gain{4.33} \\
Only TTE & \xmark & \xmark & \cmark & 27.50\loss{2.17} \\
\bottomrule
\end{tabular}
\caption{ \textbf{Component ablation on SpreadsheetBench} with the Human-Written initialization, using Qwen3.5-35B-A3B for both author and user models. }
\label{tab:ablation}
\end{table}

\begin{figure*}[t]
\centering
\includegraphics[width=\textwidth]{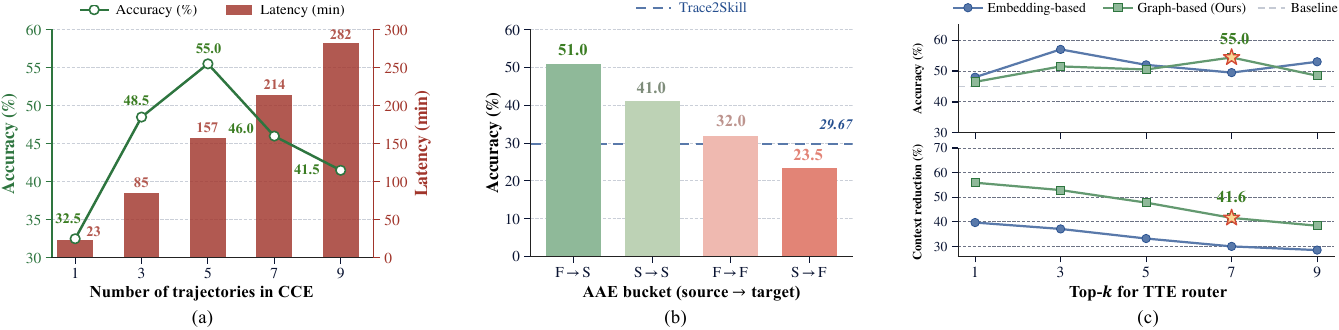}

\caption{\textbf{Additional analyses of CCE, AAE, and TTE}. (a) CCE accuracy and sampling cost across different numbers of trajectories. (b) AAE bucket accuracy by replay transition, with Trace2Skill as the reference. (c) Accuracy and context reduction of embedding and LLM graph routing across {Top-k}. All results are from Qwen3.5-35B-A3B on SpreadsheetBench.}

\label{fig:more_analysis}
\end{figure*}

\subsection{Ablation Study}
\label{sec:ablation}

Table~\ref{tab:ablation} reports the ablation of Qwen3.5-35B-A3B with Human-Written initialization, using Trace2Skill (29.67\%) as reference. The whole SkillCAT pipeline reaches 55.00\% (+25.33\%), while every leave-one-out variant declines, confirming that the modules are complementary. Specifically, without CCE, performance is 32.50\% (+2.83\%). This is because AAE can filter patches but cannot recover the same-task contrasts required for effective localized editing. Without AAE, performance falls to 26.00\% (-3.67\%), because bypassing replay validation and score-guided merging introduces error rules into the global skill. TTE can route but not correct these errors. Therefore, reliable skill content requires both contrastive evidence and patch validation. In contrast, when TTE is absent and CCE and AAE are retained, we still achieve 46.50\% (+16.83\%) performance improvement. The gap to the full pipeline shows that routing further improves execution by suppressing irrelevant content. The single-module results further reinforce this dependency. \textit{Only CCE} and \textit{Only AAE} reach 39.00\% (+9.33\%) and 34.00\% (+4.33\%), respectively, while \textit{Only TTE} reaches 27.50\% (-2.17\%). This is because routing neither creates nor validates rules, and fixed \texttt{Top-k} selection over an unrefined skill would omit useful content and select error rules. Overall, CCE improves edit evidence, AAE validates candidate edits, and TTE selectively deploys the evolved skill.

\subsection{More Analyses}
\label{sec:more_analyses}

\paragraph{How many trajectories does CCE need to form useful contrastive pairs?}
\label{sec:evidence_budget}

Figure~\ref{fig:more_analysis} (a) reports SpreadsheetBench held-out performance and CCE-stage inference time for different numbers of trajectories per task. Performance rises from 32.50\% with one trajectory to 55.00\% with five, but drops to 46.00\% and 41.50\% with seven and nine trajectories respectively, while inference time grows from 23 minutes to 282 minutes. This decline indicates that larger per-task pools include more incidental failures. Success/failure pairs built from such failures yield patches targeting unnecessary rather than task-critical behavior, adding noise to the evolved skill. Therefore, we sample five trajectories by default in our work.

\paragraph{Is it necessary to filter out low-reward skill patches using AAE?}
\label{sec:assessment_calibration}

In our SkillCAT, instead of merging all skill patches into a single skill, AAE measures the reward ($a_i$ in Eq.~\ref{eq:aae_scoring}) of each patch and selects only the high-reward ones for merging. Here, to investigate the effect of patches with different rewards, we first construct four patch buckets that contain the same number of patches but differ in reward, \textit{i.e.}, F$\xrightarrow{}$S ($a_i=3$), S$\xrightarrow{}$S ($a_i=2$), F$\xrightarrow{}$F ($a_i=1$), and S$\xrightarrow{}$F ($a_i=0$), where F and S denote \texttt{failure} and \texttt{success}, respectively. In this way, four skills are obtained and used for the subsequent TTE stage. The comparative results of Qwen3.5-35B-A3B on SpreadsheetBench are presented in Figure~\ref{fig:more_analysis}~(b), from which we observe a strong correlation between the reward and the final performance. Specifically, high-reward patches (\textit{i.e.}, F$\xrightarrow{}$S and S$\xrightarrow{}$S) yield substantial performance gains, whereas low-reward patches yield either marginal improvements (\textit{i.e.}, $F \xrightarrow{} F$) or degrade performance (\textit{i.e.}, $S \xrightarrow{} F$). Thus, we use the F$\xrightarrow{}$S and S$\xrightarrow{}$S patches (\textit{i.e.}, threshold $\theta=2$ in Eq.~\ref{eq:patch_selection}) in our work.

\paragraph{Does topology-aware routing improve accuracy while reducing inference context?}
\label{sec:routing_efficiency}

As shown in Figure~\ref{fig:more_analysis} (c), TTE successfully unleashes higher evolved-skill gains by routing and retaining only the most effective context. To filter skill content, we designed two routing mechanisms: a \texttt{graph-based router} and an \texttt{embedding-based router}. The former leverages the LLM to analyze the topological structure formed by skill content, subsequently selecting the Top-$k$ most relevant nodes, while the latter retrieves the Top-$k$ nodes based on embedding similarity computed by third-party Qwen3-Embedding-0.6B~\cite{qwen3embedding}. Both routers stay above this reference for most budgets while using fewer tokens. We chose the prompt-based method because of its competitive performance and its independence from an external embedding model, which avoids introducing additional system complexity.


\section{Conclusion}
\label{sec:conclusion}

This paper studies three limitations of existing skill self-evolution methods for LLM agents: single-trace bias, unvalidated merging, and inference-time context overload. SkillCAT addresses them by extracting same-task contrastive evidence with CCE, filtering candidate patches through source-task replay with AAE, and loading only task-relevant skill content through TTE. To assess the effectiveness of SkillCAT, we conducted evaluations on several popular agent benchmarks, such as SpreadsheetBench, WikiTableQuestions, and DocVQA, while also investigating its cross-model and OOD generalization. Under these conditions, SkillCAT outperforms comparable baselines by increasing the average score by up to 49.69 percentage points, while ablations and more analyses confirm the contribution of each module.

\bibliography{reference}

\end{document}